\UseRawInputEncoding
\documentclass{article}

\usepackage[final]{neurips_2024}

\usepackage{amssymb} 
\usepackage[utf8]{inputenc} 
\usepackage[T1]{fontenc}    
\usepackage{hyperref}       
\usepackage{url}            
\usepackage{booktabs}       
\usepackage{amsfonts}       
\usepackage{nicefrac}       
\usepackage{microtype}      
\usepackage{xcolor}         
\usepackage{amsmath}
\usepackage{wrapfig}
\usepackage{graphicx}
\usepackage{makecell}
\usepackage{pifont} 
\usepackage{multirow}

\title{GVKF: Gaussian Voxel Kernel Functions for Highly Efficient Surface Reconstruction in Open Scenes}

\author{%
  Gaochao Song\thanks{Equal contribution.} \quad Chong Cheng\footnotemark[1] \quad Hao Wang $^\dagger$\\
  AI Thrust, HKUST(GZ) \\
  \texttt{gaochaosong@hkust-gz.edu.cn} \\
  \texttt{ccheng735@connect.hkust-gz.edu.cn} \\
  \texttt{haowang@hkust-gz.edu.cn}
}

\begin{document}

\maketitle

\begin{abstract}
In this paper we present a novel method for efficient and effective 3D surface reconstruction in open scenes. Existing Neural Radiance Fields (NeRF) based works typically require extensive training and rendering time due to the adopted implicit representations. 
In contrast, 3D Gaussian splatting (3DGS) uses an explicit and discrete representation, hence the reconstructed surface is built by the huge number of Gaussian primitives, which leads to excessive memory consumption and rough surface details in sparse Gaussian areas.
To address these issues, we propose Gaussian Voxel Kernel Functions (GVKF), which establish a continuous scene representation based on discrete 3DGS through kernel regression. The GVKF integrates fast 3DGS rasterization and highly effective scene implicit representations, achieving high-fidelity open scene surface reconstruction. Experiments on challenging scene datasets demonstrate the efficiency and effectiveness of our proposed GVKF, featuring with high reconstruction quality, real-time rendering speed, significant savings in storage and training memory consumption. Project page: \url{https://3dagentworld.github.io/gvkf/}.
  
\end{abstract}

\section{Introduction}
\label{Introduction}
3D surface reconstruction in open scenes holds great significance in various practical applications, such as autonomous driving, virtual reality, urban planning and etc. However, achieving high-fidelity and efficient open scene reconstruction has been a longstanding challenge, due to the trade-off between the rendering quality and the required resources for optimization.

In pursuit of this goal, two predominant approaches are Neural Radiance Fields (NeRF) \cite{nerf,snerf,barron2021mipnerf,guo2023streetsurf, kulhanek2023tetranerf} and 3D Gaussian Splatting (3DGS) \cite{kerbl3Dgaussians,chen2024periodic, lin2024vastgaussian, yan2023streetgaussians} based methods. On one hand, NeRF-based implicit representations typically require extensive training and rendering time, which limits the practical use in large-scale scene reconstruction \cite{guo2023streetsurf, wang2023f2nerf, mi2023switchnerf, tancik2022blocknerf}. 
On the other hand, 3DGS \cite{kerbl3Dgaussians} adopts explicit representations, which enables high-quality novel view synthesis while achieving real-time rendering. This makes 3DGS more feasible for efficient scene reconstruction in the applications such as autonomous driving and virtual reality. 


Recently, there are studies using 3DGS technology for novel view synthesis and surface reconstruction in street scenes and urban environments \cite{chen2024periodic, lin2024vastgaussian, yan2023streetgaussians, matchingcubes}. For instance, SuGaR \cite{guédon2023sugar} attempts to reconstruct the 3D surfaces based on Gaussian points. However, it has been noted that overly large and sparse Gaussian points can significantly affect the geometric representations of the scene, particularly in background areas. To overcome these challenges, the 2D Gaussian Splatting (2DGS) \cite{huang20242d} proposes to use Gaussian surfaces as surfels to represent complex geometries \cite{Surfels}, thereby improving the surface reconstruction quality. Particularly, 2DGS faces challenges when processing large-scale scenes, as it requires the explicit representation of a large number of Gaussian primitives, leading to significant GPU memory consumption. 
Therefore, 2DGS still exhibits limitations in novel view synthesis capabilities and the geometric representation of large-scale scenes.

In Table \ref{fig-compare}, we summarize the comparison of 3DGS rendering and volume rendering. To fully leverage the fast rendering advantages of Gaussian alpha blending while achieving effective implicit scene representation, we propose a novel Gaussian Voxel Kernel Functions (GVKF) method. 
Firstly, GVKF utilizes voxelization to implicitly represent 3DGS, managing the growth and pruning of Gaussian splats. This approach retains the expressive power of explicit Gaussian splats while enabling efficient management of these splats. 
Secondly, we carefully analyze the intrinsic connection between Gaussian splatting alpha blending rendering and traditional volume rendering from a mathematical perspective. We establish a 3DGS-based method to represent continuous scene opacity density fields through kernel regression. This makes it possible for discrete Gaussians to represent continuous scenes. By replacing the discrete opacity values in original 3DGS rendering pipeline (which can be viewed as collapsed kernel functions) with Gaussian kernel functions, we maintain the advantages of the original 3DGS alpha blending while optimizing the representation of continuous scenes. Moreover, we demonstrate that our proposed rendering method is mathematically consistent with traditional volume rendering. 
Thirdly, based on our constructed scene opacity representation, which is also known as the scene opacity field, we derive the bidirectional mapping relationship between opacity and the scene surface. This enables direct mesh extraction for scene surface. In summary, our contributions are as follows:
\begin{itemize}
    \item We propose GVKF, an implicit continuous scene reconstruction method that integrates the effectiveness of implicit representation with the fast rasterization advantages of Gaussian Splatting, without the need for computationally intensive volume rendering.
    \item Based on GVKF, we further propose implicit representation of the scene surface, achieving efficient and high-quality scene surface reconstruction.
    \item Experiments demonstrate the usefulness of GVKF in open scenes, showcasing high-quality surface reconstruction accuracy, real-time rendering speeds, and significant savings in storage and memory consumption.
\end{itemize}

\begin{table}
\centering
\small
\caption{Comparison of 3DGS rendering and volume rendering methods.}
\resizebox{\textwidth}{!}{
\begin{tabular}{l|l|l|p{4cm}}
\toprule
\textbf{Method} & \textbf{Math Expression} & \textbf{Pros} & \textbf{Cons} \\
\midrule
3DGS Rendering & Discrete summation & Fast rendering & High Mem consumption, Hard to fit 3D continuous surface \\
\midrule
Volume Rendering & Continuous integration & Better 3D surface representation & Low rendering speed due to continuous sampling \\
\bottomrule
\end{tabular}
}
\label{fig-compare}
\vspace{-15pt}
\end{table}

\vspace{-7pt}
\section{Related Works}
\label{Related Works}
\subsection{Novel View Synthesis}
The introduction of Neural Radiance Fields (NeRF) \cite{nerf} has significantly advanced the development of 3D reconstruction and novel view synthesis. NeRF employs volumetric rendering techniques to intricately simulate the geometric structure of scenes and viewpoint-dependent characteristics, thereby considerably enhancing the quality of image rendering. Following NeRF, variants such as Mip-NeRF \cite{barron2021mipnerf} and Zip-NeRF \cite{barron2023zipnerf} have addressed the aliasing issues during rendering. Additionally, UC-NeRF \cite{ucnerf}, designed for outdoor scenes, enhances image consistency through color correction and pose refinement. InstantNGP \cite{instantngp} accelerates training and improves rendering efficiency by optimizing subvolume processing with grid pyramid techniques. Meanwhile, other feature grid-based scene representation methods \cite{chen2022tensorf, liu2021neural, sun2022direct, chen2023neurbf, yu2021plenoctrees} have been extensively explored to enhance the training capability and expressiveness of models. Recently, 3D Gaussian Splatting (3DGS) \cite{kerbl3Dgaussians} effectively represents complex scenes using 3D Gaussian points, significantly boosting the efficiency of real-time high-resolution image rendering while maintaining rendering quality. Further research efforts like Scaffold-GS and Octree-GS \cite{lu2023scaffoldgs, ren2024octreegs} have attempted more effective methods to organize and manage Gaussian points, which helps reduce memory usage and speed up training.

\vspace{-7pt}
\subsection{Surface Reconstruction}
Traditional isosurface extraction, relying on density thresholds, often struggles with fine details due to resolution and noise constraints. Recent studies propose more complex representation methods \cite{zhang2021ners}. For instance, NeuS \cite{wang2023neus} uses MLP networks for occupancy grids or SDF, improving reconstruction accuracy and noise reduction \cite{oechsle2021unisurf, NEURIPS2021_25e2a30f, li2023neuralangelo, darmon2022improving, yang2022neumesh}. Techniques like BakedSDF \cite{yariv2023bakedsdf} translate the optimization of NeRF or neural SDFs into 3D meshes, enhancing features through high-resolution grids but increasing computational load. NeuS2 \cite{wang2023neus2} introduces a novel formula for second-order derivatives with multi-resolution hash encoding and CUDA-based MLP technology, significantly reducing training time. StreetSurf \cite{guo2023streetsurf} optimizes SDF mappings in open scenes and decouples static and dynamic objects. Despite advancements, NeRF-based methods still need optimization for processing speed and real-time rendering.

3DGS has gained attention for its high-quality scene reconstruction and rapid processing capabilities \cite{kerbl3Dgaussians}. 3DGS uses multiple 3D Gaussian distributions with anisotropic covariance for precise control over scene attributes \cite{964490, Kopanas_2022}. This technology enhances surface reconstruction methods like SuGaR \cite{guédon2023sugar}, which employs Poisson surface reconstruction for fast and accurate mesh extraction. However, irregular Gaussian sphere distribution affects surface quality. To improve this, 2DGS \cite{huang20242d} uses 2D Gaussian planes for better surface conformity and TSDF for accurate reconstruction, though it may cause surface fragmentation. GOF \cite{yu2024gaussian} directly extracts surfaces using opacity thresholds and tetrahedral mesh extraction but is limited by high VRAM requirements. GSDF \cite{yu2024gsdf} combines 3DGS with a NeuS-like SDF branch for optimized rendering and reconstruction, increasing training time. Despite their potential, 3DGS-based methods face challenges like managing Gaussian points, high VRAM consumption, and degraded rendering quality. 



\section{Methods}
\label{Methods}
\begin{figure} 
\centering
\includegraphics[width=\textwidth]{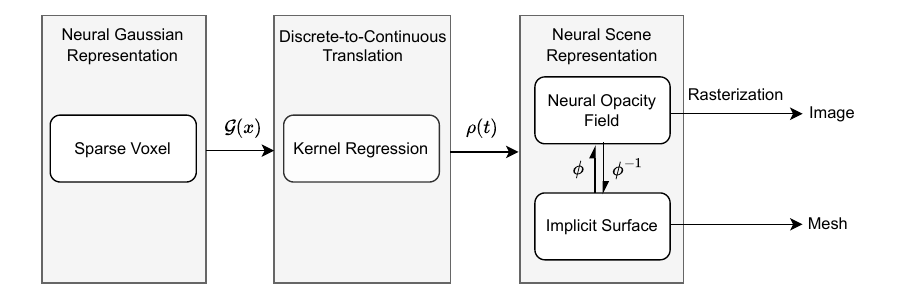} 
\caption{Framework of Gaussian Voxel Kernel Functions (GVKF) for scene representation. In this framework, discrete Gaussian primitives $\mathcal{G}$ represent continuous opacity density $\rho(t)$ on the ray via kernel regression. After slightly modifying the rasterization pipeline, the kernel function can be integrated into alpha blending rasterization without introducing dense points sampling. Additionally, we directly define the mapping relationship between the neural opacity field and the implicit surface.}
\label{main-main}
\vspace{-7pt}
\end{figure}

As shown in Fig. \ref{main-main}, we first introduce the implicit neural 3DGS primitives representation based on a sparse voxel grid, which offers highly efficient storage management and the fitting power of neural networks. Secondly, we present our GVKF-based continuous scene representation, to explain its rationale, we have analyzed its intrinsic connections with Gaussian alpha blending \cite{kerbl3Dgaussians} and traditional volume rendering \cite{nerf} from a statistical analysis perspective. Finally, we describe the relationship between the proposed continuous scene representation (a neural opacity field) and implicit surface, and derive an explicit mapping function for mesh reconstruction.

\subsection{Voxel Gaussian Representation}
To achieve orderly 3DGS management while minimizing the explicit expression of them to save training storage consumption, we use a spatial sparse voxel grid to manage Gaussian primitives. During the initialization phase, the sparse grid is generated from the downsampled SfM point clouds and dynamically grows or being eliminated during training. Each sparse grid is allowed to generate up to $m$ Gaussian primitives, and all these primitives are limited to a small range of space centered at the voxel grid.

\textbf{Gaussian Generation}
For a particular 3D Gaussian expression, five attributes are required: $p \in \mathbb{R}^3$ (position), $\alpha \in \mathbb{R}$ (opacity), $R \in \mathbb{R}^{3 \times 3}$ (rotation matrix), $s \in \mathbb{R}^3$ (scaling), and $c \in \mathbb{R}^3$ (color). Then, a Gaussian $\mathcal{G}(x)$ can be generated as:
\begin{equation}\label{origin3dgs}
    \mathcal{G}(x)=\alpha \cdot e^{-\frac{1}{2}(x-p)^T\sum\nolimits^{-1}(x-p)},
\end{equation}
where $\sum \in \mathbb{R}^{3\times 3}$ is covariance matrix defined as $\sum = Rss^TR^T$. $c$ is calculated via SH coefficients and camera direction. Different from traditional 3DGS\cite{kerbl3Dgaussians} that treats them as explicit optimizable tensors, we decode them from a feature vector $\mathcal{F} \in \mathbb{R}^d$ via several MLPs:
\begin{equation}
    \alpha=\text{MLP}_\alpha(\mathcal{F},\text{camera}), R=\text{MLP}_R(\mathcal{F}), s=\text{MLP}_s(\mathcal{F}), c=\text{MLP}_c(\mathcal{F}, \text{camera}).
\end{equation}
For alpha and color MLPs, the view camera and feature vector $\mathcal{F}$ are inputs, facilitating view-dependent fitting. Relative coordinates of Gaussians to the parent voxel center are stored with $\mathcal{F}$, compressing explicit Gaussian components and leveraging the MLP’s fitting capacity. Gaussians are dynamically generated each iteration and recycled post-update, reducing memory usage.

\textbf{Voxel Registration}.
To control Gaussian numbers in large open scene, we eschew the traditional adaptive density control strategy, adopting a method inspired by scaffold-Gaussian \cite{lu2023scaffoldgs} and Octree-Gaussian \cite{ren2024octreegs}. The voxel registration is based on gradient accumulation. After each iteration, gradients from 3DGS are recorded and accumulated in their respective voxels, denoted as $\nabla$. Voxels where $\nabla$ exceeds a set threshold are subdivided into eight subvoxels to increase grid resolution, continuing until the maximum depth is reached. Additionally, less frequently used voxels are discarded after a specified period.

\subsection{Neural Opacity Field of 3DGS}

Since 3DGS rasterization rendering and traditional volume rendering share some overlapping concepts, in this section, we sort them out and introduce our method from a statistical perspective while avoiding introducing redundant mathematical symbols.

\textbf{Continuous Scene Description}. We define $\rho(t): [0,+\infty] \rightarrow [0,1]$ as the opacity density function, which measures the probability of a ray encountering a particle at position $t$. We define $\mathcal{T}(t): [0,+\infty] \rightarrow [0,1]$ as the transmission function, which measures the probability that a ray has not encountered any particles from its origin to point $t$. Considering the probability that a ray does not encounter any particles at time step $t+dt$, denoted as $\mathcal{T}(t+dt)$, it is evident that $\mathcal{T}(t+dt)=\mathcal{T}(t)(1-\rho(t)dt)$. Solving this differential equation, we obtain the relationship between $\mathcal{T}(t)$ and $\rho(t)$:
\begin{equation}
    \mathcal{T}(t)=\exp(-\int_{0}^t\rho(t)dt).
\end{equation}
Therefore, we obtain the cumulative distribution function (CDF) of the probability that a ray \textbf{hits} a particle over the interval $[0, t]$: $\Phi(t) = 1 - \mathcal{T}(t)$, with the corresponding probability density function (PDF) being $\Phi^\prime(t) = \mathcal{T}(t) \cdot \rho(t)$. From the perspective of volume rendering, this PDF is used as the probability of the appearance of color along the ray, ultimately taking the mathematical expectation of the color as the ray color:
\begin{equation}\label{eq-vol}
    C=\int_{0}^{B} \mathcal{T}(t) \cdot \rho(t) \cdot c(t)dt + \mathcal{T}(B) \cdot c_{bg}.
\end{equation}
The discrete formulation of volume rendering Eq. \ref{eq-vol} is:
\begin{equation}\label{eq-vol-disc}
    C=\sum_{i=1}^{N}T_i \cdot \alpha_i \cdot c_i, \quad   \alpha_i=(1-\exp(-\sigma_i\delta_i)), \quad T_i=\prod_{j=1}^{i-1}(1-\alpha_j)
\end{equation}

where opacity $\alpha_i$ represents the accumulated result in a sampling interval $\delta_i$ of volume density $\sigma_i$, hence the value of $N$ does not influence the result as long as $\alpha_i$ is adapted enough. Based on the similar idea of volume rendering, the PDF $\Phi^\prime(t)=\mathcal{T}(t) \cdot \rho(t)$ can also be reasonably considered as the probability of the appearance of a surface along the ray, where the place with the highest probability density is most likely to have a surface. Correspondingly, on the CDF $\Phi(t)$, this is the place where the derivative is the largest. In this paper, we use the CDF $\Phi(t)$ to describe continuous scenes based on the camera rays, to facilitate integration with the 3DGS rasterization rendering pipeline. In section \ref{sec-mapping} , we will prove that under the 3DGS representation, the place where the derivative of CDF is the largest is not actually the surface, so a method to locate the surface will be introduced.

\begin{wrapfigure}{r}{0.48\textwidth} 
\vspace{-15pt}
    \centering
    \includegraphics[width=0.48\textwidth]{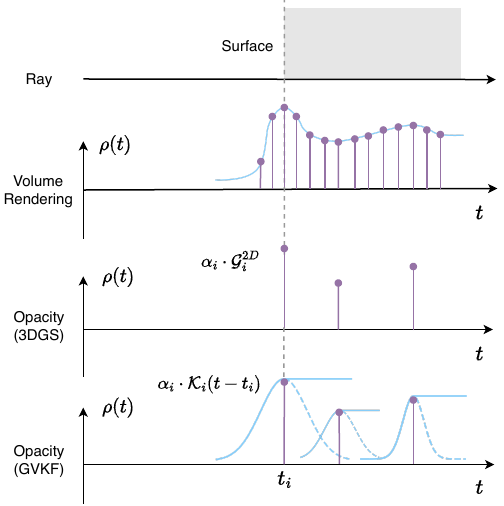} 
    \caption{Comparison of Volume Rendering, 3D Gaussian Splatting with Alpha Blending, and GVKF Rendering.}
    \label{fig:comp-rendering}
    \vspace{-20pt}
\end{wrapfigure}

\textbf{Kernel Regression of 3DGS}. 

In implicit scene representation methods based on volume rendering, the continuous opacity density function is directly predicted by a MLP\cite{nerf}. In our approach, the continuous opacity density function $\rho(t)$ is fitted through kernel regression via discrete Gaussian primitives after a differentiable transformation: $\mathcal{G}_i(x, y, z) \rightarrow \mathcal{K}_i(t-t_i)$ from 3D to 1D. The transformation consists of three steps: (1). The Gaussian primitives that the ray passes through are selected as kernel functions. (2). According to the Ray-Gaussian Intersection method \cite{intersec2}, the ray is transformed into the local coordinate system of each 3DGS to obtain the 1D probability density alone the ray. Here, the peak of the 1D probability density, denoted as $t_i$, is defined as the Ray-Gaussian Intersection \cite{intersec1,intersec2,yu2024gaussian}, indicating that the 3DGS has the greatest influence at this point alone the ray. (3). To integrate with regularization methods \cite{huang20242d, yu2024gaussian}, we assume that each 3DGS fits the surface of the object. Therefore, after $t_i$, the probability density continues to remain at its maximum value, indicating that the object is solid. Without loss of generality, the opacity density $\rho(t)$ on camera ray can be expressed as:
\begin{equation}\label{eq-kernel}
    \rho(t)=\sum_i^{N} \alpha_i \cdot \mathcal{K}_i(t-t_i), \quad
\mathcal{K}_i(t)=
\begin{cases} 
\exp(-k_i \cdot t^2) & t < 0 \\
1 & t \geq 0,
\end{cases}
\end{equation}
where $N$ represents the number of activated kernel functions along the ray, and $k_i$ represents the summarized transform of Ray-Gaussian transform (See Appendix \ref{appendix:raygaussian} for details). $\alpha_i$ represents the coefficient for each kernel function.\footnote{Specifically, $\alpha_i=\beta_i \frac{\sqrt{k_i}}{\sqrt{\pi}}$, where $\beta_i$ is a constant value representing opacity related to Gaussian primitives.}

\textbf{Rendering}. As for traditional 3DGS rasterization, the pixel color is rendered through alpha blending on $N$ 3DGS being passed through by the ray:
\begin{equation}\label{eq-gsrender}
    C=\sum_{i=1}^N c_i \cdot \alpha_i \cdot \mathcal{G}_i^{2D} \prod_{j=1}^{i-1}(1-\alpha_j \cdot \mathcal{G}_j^{2D})
\end{equation}
In this scenario, $\alpha_i$ is constant value representing the opacity of Gaussians. This point-based rendering is coherent with Eq. \ref{eq-vol-disc}, with extremely sparse sampling points to simulate dense volume rendering. However, it is impossible to recover continuous opacity density alone the ray from such a rendering equation, as illustrated in row-3 of Fig. \ref{fig:comp-rendering}. This is because the third row of the covariance matrix of 3DGS is discarded, and it is directly projected onto a 2D plane to evaluate the impact on the opacity of points along the ray. From the perspective of Eq. \ref{eq-kernel}, this means that along the ray, the influence range of all $N$ kernel functions that intersect with the ray collapses to an infinitesimally value, making it impossible to recover a continuous opacity density field. To solve this, Eq. \ref{eq-gsrender} can be modified to: \footnote{This rendering form is firstly proposed by GOF \cite{yu2024gaussian}}
\begin{equation}\label{eq-gsrender2}
    C=\sum_{i=1}^N c_i \cdot \alpha_i \cdot \mathcal{K}_i(0) \prod_{j=1}^{i-1}(1-\alpha_j \cdot \mathcal{K}_j(0))
\end{equation}
This equation 
will not affect the goal of 3DGS rendering: to approximate traditional volume rendering using sparse sampling points. And it allows for broadening the collapsed kernel functions to fit the continuous opacity function of the scene.

\textbf{Scene Representation}. Based on the discussion at the beginning, the scene surface can be described via CDF $\Phi(t)$ on the ray in a continuous way, which can be calculated like Eq. \ref{eq-gsrender2} (removing color):
\begin{equation}\label{eq-scene}
    \Phi(t)=\sum_{i=1}^{N}\alpha_i \cdot \mathcal{K}_i(t-t_i) \prod_{j=1}^{i-1}(1-\alpha_j \cdot \mathcal{K}_j(t-t_j))
\end{equation}

\begin{wrapfigure}{r}{0.4\textwidth} 
\vspace{-15pt}
    \centering
    \includegraphics[width=0.4\textwidth]{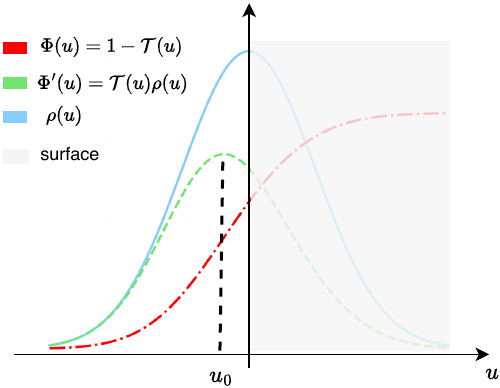} 
    \caption{Illustration of functions $\Phi(u),\Phi^\prime(u),\rho(u)$.}
    \label{ii}
\end{wrapfigure}
\subsection{Implicit Surface Mapping}
\label{sec-mapping}
This implicit opacity field (denoted as neural opacity field since it is represented by neural Gaussians) measures the CDF of the probability that a ray hits solid scene surface. In the next section, we introduce the mapping of $\Phi(t)$ to implicit surface.

We represent implicit surface with signed distance function (SDF), denoted as function $D(t)$ on the camera ray. To recover $D(t)$ of given $\Phi(t)$ that is calculated from well trained 3DGS, we firstly study the reverse mapping problem: $\phi : D(t) \rightarrow \Phi(t)$

\textbf{Opacity Density Near the Surface}.
To ensure that the 3DGS aligns with the object's surface and thus reflects the object's shape, depth distortion regularization \cite{huang20242d, yu2024gaussian} is introduced during the Gaussian training process. This encourages the distribution of 3DGS along the ray to aggregate together, causing the peak of the kernel functions to coincide with the object's surface. In the next discussion, the coordinate of object surface on the ray is assumed as $t^*$ with $D(t^*)=0$. Considering $\rho(t)$ at the interval $t\in [0,t^*]$, we have:
\begin{equation}
    \rho(t)=\rho(t^*-D(t))=\sum_{j=1}^{M} \alpha_j \cdot \exp(-k_j \cdot D(t)^2), \quad 0 \leq t \leq t^*
\end{equation}

Where $M$ represents the number of Gaussian kernels concentrated on the surface. To facilitate calculations, we convert the opacity density system to the SDF coordinate system, with $t^*$ as the origin and letting $u=-D(t)$, as illustrated in Fig. \ref{ii}, we have:
\begin{align}
    \Phi^{\prime} (u) &= \mathcal{T}(u) \cdot \rho(u) = \exp (-\int_{-t^*}^u \rho(w)dw ) \cdot \rho(u)
\\
    \Phi^{\prime\prime}(u) &= -\rho^2(u) \cdot \exp (-\int_{-t^*}^u \rho(w)dw ) + \rho^\prime(u) \cdot \exp (-\int_{-t^*}^u \rho(w)dw ) \notag \\
    &= [-\rho^2(u)+ \rho^\prime(u)]\exp (-\int_{-t^*}^u \rho(w)dw )
\end{align}
where $\rho(u) \sim \mathcal{N} (0, \sigma^2), \sigma^2=\sum_{i=1}^M \frac{1}{2\pi \alpha_i^2}$, which can be directly derived from the additive property of the normal distribution.
Then letting $h(u)=-\rho^2(u)+ \rho^\prime(u)$, we have:
\begin{align}\label{eq-all}
    h(u) &= -\rho^2(u)+ \rho^\prime(u) \notag \\
    &= -\rho(u)[\rho(u)+\frac{u}{\sigma^2}]
\end{align}
It is easy to prove that $h(u)$ crosses a unique zero point $u_0$ from top to bottom on the u-axis, and $u_0 < 0$. This means that the peak of $\Phi^\prime(u)$ will appear before the surface, so it is not reasonable to simply determine the actual intersection point of the light ray with the surface by directly evaluating the peak of $\Phi^\prime(u)$. To locate the accurate surface, a transcendental equation of $u$ is needed to be solved to get $u_0$:
\begin{equation}
    \rho(u)=-\frac{u}{\sigma^2}, \quad \rho(u)=\frac{1}{\sqrt{2\pi}\sigma} \exp(-\frac{u^2}{2\sigma^2}), \quad \sigma^2=\sum_{i=1}^M \frac{1}{2\pi \alpha_i^2}
\end{equation}

\begin{wrapfigure}{r}{0.45\textwidth} 
    \vspace{-10pt}
    \centering
    \includegraphics[width=0.45\textwidth]{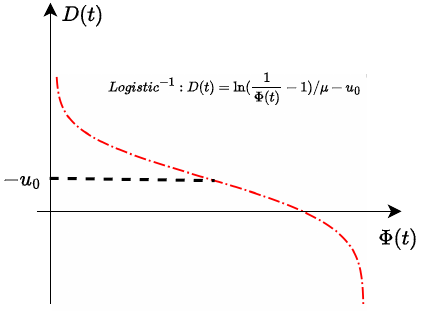} 
    \caption{Illustration of opacity to SDF mapping.(Eq. \ref{eq-map})}
    \label{mapppp}
    \vspace{-10pt}
\end{wrapfigure}

It is impossible to directly get the analytical solution, however, numerical computation methods can be applied to solve $u_0$. This may require some extra time for computation.

\begin{figure}
\centering
  \includegraphics[width=1\linewidth]{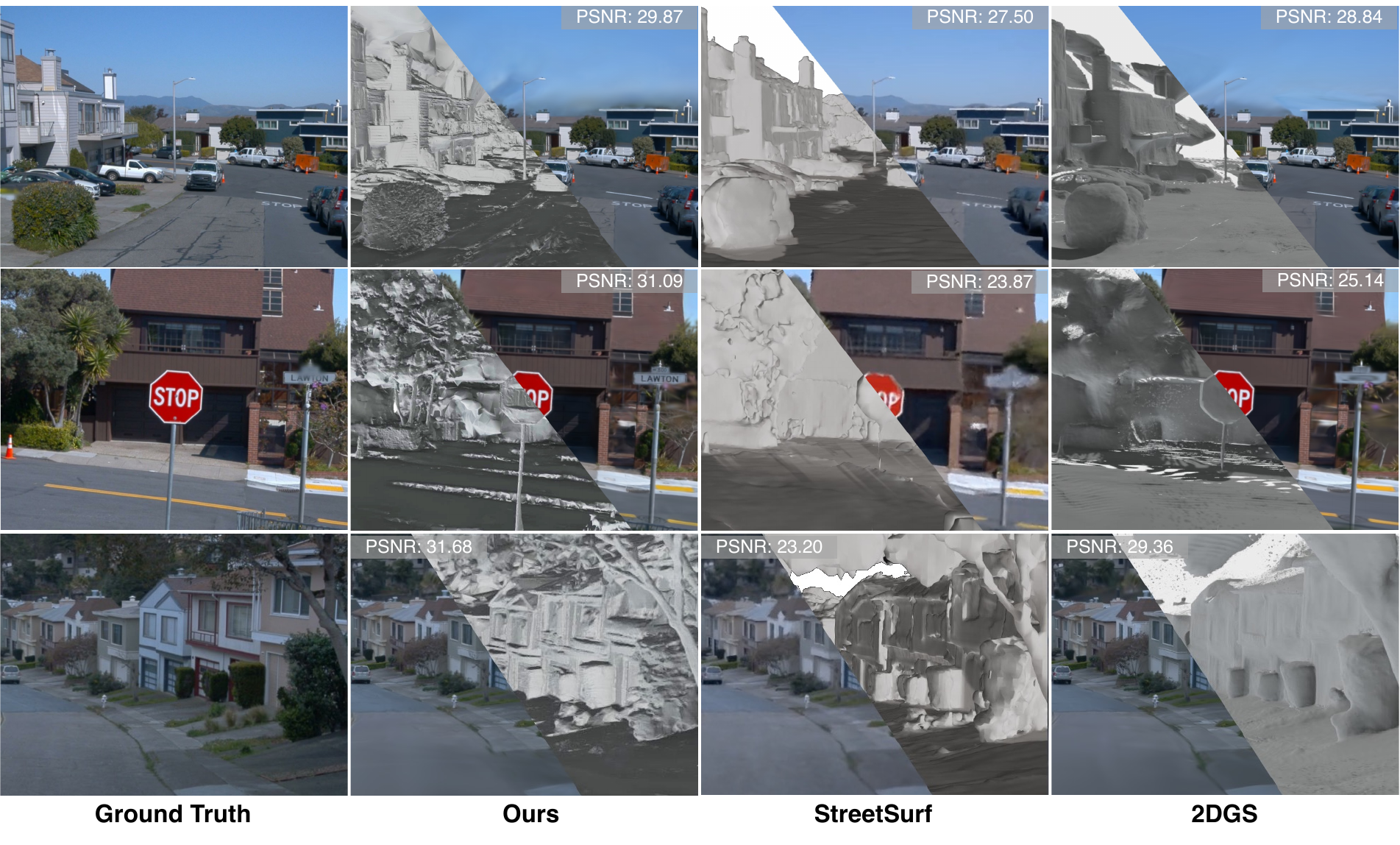}
  \vspace{-5pt}
  \caption{Qualitative comparison of novel view synthesis and surface reconstruction on the Waymo Open Dataset~\cite{Sun}, with each subplot annotated with PSNR values to quantify image quality. Our method shows higher geometric precision and detail, validating its efficiency and superiority in processing open scenes, especially in geometric accuracy and detail reproduction.}
  \label{fig:waymo}
\vspace{-10pt}
\end{figure}

\textbf{Mapping from Opacity to Surface}.

Based on the analysis above, we can always have exact number of $u_0$ via numerical computation method. However, it is hard to find out the inverse function of $\Phi(u)$ for directly building the mapping of $\Phi(t)$ to $D(t)$. For the balance of surface smoothing while reducing the indelible error, we represent mapping relationship of $ u \rightarrow \Phi(u)$ via Logistic Function as follows:

\begin{equation}
\Phi(u) = \frac{1}{1+\exp(-\mu(u-u_0))}    
\end{equation}

where $\mu$ represents the smooth factor. We choose Logistic Function because of its formal is concise and shares similar shape of $\Phi(u)$. More importantly, it only has one inflection point at $(0,0.5)$, which can be used to simulate the inflection point of $\Phi(u)$ after translation.  Finally, we represent implicit SDF function via Inverse function transformation of Logistic Function, as shown in Fig. \ref{mapppp}:
\begin{equation} \label{eq-map}
    D(t)=\ln(\frac{1}{\Phi(t)}-1)/\mu - u_0
\end{equation}

\section{Experiments}
\label{Experiments}
\subsection{Experimental Settings}

\begin{figure}
\centering
  \includegraphics[width=1\linewidth]{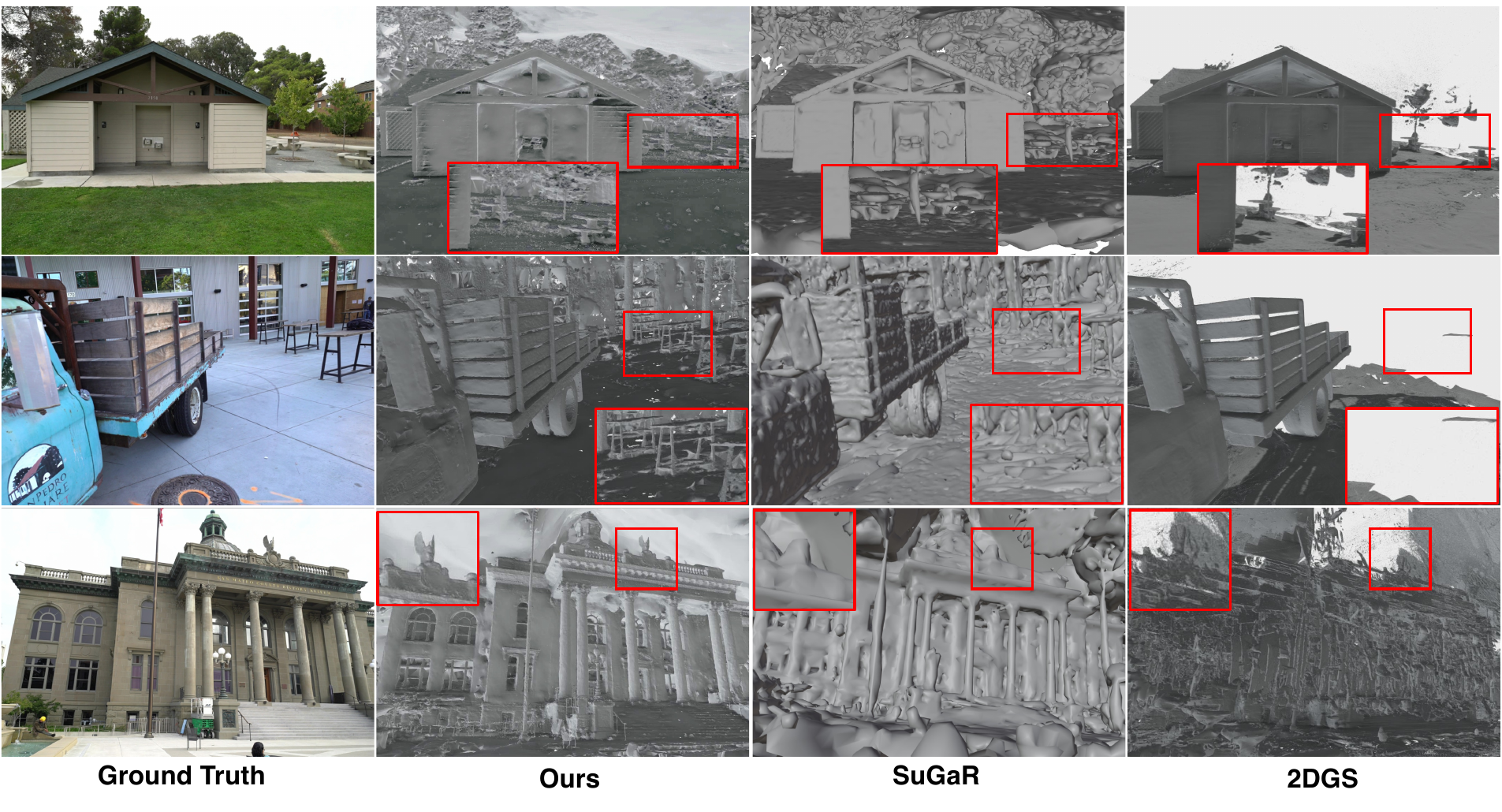}
  \caption{Qualitative comparison on the Tanks and Temples dataset \cite{tnt} shows that our method excels in reconstructing complex backgrounds with high geometric granularity. In contrast, 2DGS often results in fragmented backgrounds, while SuGaR displays uneven spherical shapes, affecting both visual and geometric quality.}
  \label{fig:tnt}
\end{figure}


\textbf{Datasets.}
To assess our method's performance against baseline methods in open scenes, we used three datasets. We first experimented with the Waymo Open Scene dataset~\cite{Sun}, using three cameras per scene from five available, each scene containing about 600 images. We employed LiDAR point clouds to evaluate reconstruction quality, although LiDAR data was not used as training input. We also tested on the Tank and Temple dataset~\cite{tnt}, which includes trajectories and ground truth for six selected scenes. Lastly, we evaluated the Mip-NeRF 360 dataset~\cite{barron2021mipnerf}; due to the absence of ground truth, our focus was on novel view synthesis to demonstrate our method's efficacy in this aspect.
\begin{table}
\centering
\caption{Quantitative evaluation of novel view synthesis and surface reconstruction on the Waymo Open Scene dataset~\cite{Sun}. Using LiDAR data as ground truth, we calculated Chamfer Distance (C-D) values for reconstruction accuracy. Our method performs excellently in both novel view synthesis and surface reconstruction, outperforming other methods in Gaussian point usage, VRAM occupancy, and real-time rendering.}

\begin{tabular}{@{}lcccccc@{}}
\toprule
\textbf{Method}    & PSNR $\uparrow$  & C-D $\downarrow$ & MB (Storage) $\downarrow$ & GB (GPU) $\downarrow$ & FPS $\uparrow$ & Training Time $\downarrow$\\ \midrule
NeuS       & 13.24     & \textbf{0.76}    & 170      & 31     & $\sim$ 0.1   &  5 h \\
$\text{F}^2$-NeRF   & 24.70     & 886.77    & 130   & 24     & $\sim$ 0.1  &  0.8 h  \\
StreetSurf & 27.12     & 1.02    & 540    & 22    & $\sim$ 0.1   & 1.5 h \\
3DGS       & 27.99 & 3.57    & 230        & 23    & \textbf{63}   & 0.75 h\\
SuGaR      & 23.71     &  3.08   & 228    & 33        & 56   & 1.5 h \\
2DGS       & 28.51     &   1.67  & 238     & 23       & 51   & \textbf{ 0.7 h} \\
GVKF (Ours)        & \textbf{30.24}     & 1.57    & \textbf{30}     & \textbf{14}       & 32 &  1.5 h \\
\bottomrule
\end{tabular}
\label{tab:waymo}
\end{table}

\begin{table}
\centering
\caption{Quantitative evaluation on the Tanks and Temples dataset~\cite{tnt} using F1 scores and training time as metrics. Our method outperforms all existing explicit methods in F1 scores and is comparable to implicit methods in reconstruction accuracy, with significantly reduced training time. These results highlight our method's efficiency and accuracy. Comparation of concurrent work GOF \cite{yu2024gaussian} is presented in Appendix \ref{appendix:comptogof}.}
\begin{tabular}{@{}lcccccccc@{}}
\toprule
 \multirow{2}{*}{\textbf{Method}}             & \multicolumn{3}{c}{Implicit}      & \multicolumn{3}{c}{Explicit}    &       \\
              \cmidrule(lr){2-4} \cmidrule(lr){5-7}
              & NeuS & Geo-NeuS & Neuralangelo & SuGaR & 3DGS & 2DGS & Ours \\ \midrule
Barn          & 0.29 & 0.33     & \textbf{0.70}         & 0.14  & 0.13 & 0.36 & \textbf{0.40} \\
Caterpillar   & 0.29 & 0.26     & \textbf{0.36}         & 0.16  & 0.08 & 0.23 & \textbf{0.34} \\
Courthouse    & 0.17 & 0.12     & \textbf{0.28}         & 0.08  & 0.09 & 0.13 & \textbf{0.25} \\
Ignatius      & 0.83 & 0.72     & \textbf{0.89}         & 0.33  & 0.04 & 0.44 & \textbf{0.51} \\
Meetingroom   & 0.24 & 0.20     & \textbf{0.32}         & 0.15  & 0.01 & 0.16 & \textbf{0.23} \\
Truck         & 0.45 & 0.45     & \textbf{0.48}         & 0.26  & 0.19 & 0.26 & \textbf{0.40} \\
\midrule
Mean          & 0.38 & 0.35     & \textbf{0.50}         & 0.19  & 0.09 & 0.30 & \textbf{0.36} \\
Time          & \textgreater 24 h & \textgreater 24 h & \textgreater 24 h & \textgreater 1 h   & \textbf{$\sim$15 min} & $\sim$30 min & $  \sim $1.5 h \\
\bottomrule
\label{tab:tnt}
\end{tabular}
\vspace{-10pt}
\end{table}

\textbf{Baselines.} In terms of surface reconstruction, we presented the results on the Waymo dataset~\cite{Sun} in tables \ref{tab:waymo} and figures \ref{fig:waymo}, comparing state-of-the-art implicit methods (such as NeuS \cite{wang2023neus}, $\text{F}^2$-NeRF \cite{wang2023f2nerf}, StreetSurf \cite{guo2023streetsurf}) and explicit methods (such as 3DGS \cite{kerbl3Dgaussians}, SuGaR \cite{guédon2023sugar}, 2DGS \cite{huang20242d}). We utilized PSNR to evaluate the results of novel view synthesis and Chamfer distance to measure reconstruction accuracy, while also recording training time, VRAM usage, and the size of the Gaussian point files post-training.  Additionally, as shown in Table \ref{tab:tnt} and Figure \ref{fig:tnt}, we conducted comparisons on the Tank and Temple dataset with implicit methods (such as NeuS \cite{wang2023neus}, Geo-NeuS \cite{fu2022geoneus}, Neuralangelo \cite{li2023neuralangelo}) and explicit methods (such as 3DGS \cite{kerbl3Dgaussians}, SuGaR \cite{guédon2023sugar}, 2DGS \cite{huang20242d}). We used official scripts to evaluate F1 scores. For novel view synthesis, we compared various advanced methods on the Mip-NeRF 360 dataset, including NeRF \cite{nerf}, Deep Blending \cite{deepblend}, Instant NGP \cite{instantngp}, MERF \cite{merf}, Mip-NeRF 360 \cite{barron2021mipnerf}, BakedSDF \cite{yariv2023bakedsdf}, 3DGS \cite{kerbl3Dgaussians}, SuGaR \cite{guédon2023sugar}, and 2DGS \cite{huang20242d}. We use evaluation metrics such as PSNR, SSIM, and LPIPS.

\begin{wraptable}{r}{0.5\textwidth} 
\vspace{-20pt}
\centering
\caption{Quantitative evaluation on the Mip-NeRF 360~\cite{barron2021mipnerf} outdoor scene dataset is presented. Since the dataset lacks ground truth for surface reconstruction, we assessed the results of novel view synthesis.}
\begin{tabular}{@{}lccc@{}}
\toprule
Method          & PSNR $\uparrow$ & SSIM $\uparrow$ & LPIPS $\downarrow$\\ \midrule
NeRF            & 21.46 & 0.458 & 0.515 \\
Deep Blending   & 21.54 & 0.524 & 0.364 \\
Instant NGP      & 22.90 & 0.566 & 0.371 \\
MERF            & 23.19 & 0.616 & 0.343 \\
Mip-NeRF 360      & 24.47 & 0.691 & 0.283 \\
BakedSDF        & 22.47 & 0.585 & 0.349 \\
3DGS            & 24.24 & 0.705 & 0.283 \\
SuGaR           & 22.76 & 0.631 & 0.349 \\
2DGS            & 24.33 & 0.709 & 0.284 \\
GVKF (Ours)     & \textbf{25.47} & \textbf{0.757} & \textbf{0.240} \\ \bottomrule
\end{tabular}
\vspace{-30pt}
\label{table:mip}
\end{wraptable}

\textbf{Implementation Details}
Our method modifies the representation of 3DGS and slightly adjusts the opacity weights in the rendering pipeline using Gaussian kernel functions. This ensures compatibility with other components of Gaussian rasterization rendering. Similarly, we employ the same L1 loss and D-SSIM loss as 3DGS to supervise color loss, and we use the same Gaussian regularization term as 2DGS and GOF to promote alignment between the Gaussians and the surface. After training, the SDF field of the scene can be directly extracted based on Eq. \ref{eq-map} and exported to a mesh with the MC\cite{mcal}/MT\cite{mt1,mt2} algorithm. To export complete sky and background, the modified MT algorithm in GOF\cite{yu2024gaussian} is used.

\vspace{-5pt}
\subsection{Analysis}
Figure \ref{fig:waymo} demonstrates the superiority of our method in capturing detailed features of roadside houses, bushes, and other objects. In contrast, the 2DGS method produced more holes and fragmentation, while the StreetSurf method lost some critical geometric features. The results in Table \ref{tab:waymo} indicate that our method surpasses other methods in terms of view synthesis and reconstruction accuracy, and it requires fewer Gaussian points and VRAM for large-scale scene reconstructions. Figure \ref{fig:tnt} highlights our method's excellent performance in scene restoration. The SuGaR method generated excessive irregular protrusions, and 2DGS exhibited more fragmentation and floating debris. According to the results in Table \ref{tab:tnt}, our method outperforms all explicit methods and achieves comparable reconstruction results to implicit methods, while maintaining equivalent GPU time usage. The results in Table \ref{table:mip} confirm that our method leads in novel view synthesis across all compared methodologies.

\subsection{Ablation Study}

In this section, we evaluate the impact of varying voxel grid sizes on the neural Gaussians by conducting experiments with the Waymo datasets. We selected voxel sizes of ${1, 0.1, 0.01, 0.001}$, as presented in Table \ref{tab:abl-1}. When the voxel size is too large, the sparse neural Gaussians fail to learn the scene representation and return NaN errors. As the number of voxels increases, more Gaussians are generated for scene representation, thereby enhancing the quality of novel view synthesis. However, the improvements plateau when the voxel size is reduced to 0.001, which also requires more training time and becomes impractical. Therefore, we set the voxel size to 0.01 to balance training time and rendering quality. 


\begin{wraptable}{r}{0.5\textwidth}
\vspace{-15pt}
\setlength{\tabcolsep}{3pt} 
\caption{Influence of different voxel size.}
\small
\begin{tabular}{ccccc}
\toprule
Voxel Size & Initial voxels & Final voxels & PSNR  & Time \\ \midrule
1          & $\sim$ 2 k          & -              & -     & -    \\
0.1        & $\sim$ 80 k       & $\sim$ 110 k       & 29.34 & 1.2 h \\
0.01       & $\sim$ 90 k      & $\sim$ 1100 k    & 30.24 & 1.5 h \\
0.001      & $\sim$ 100 k      & $\sim$ 1100 k     & 30.29 & 4 h   \\ \bottomrule
\end{tabular}
\label{tab:abl-1}
\vspace{-15pt}
\end{wraptable}

We further conducted ablation study on the Tanks and Temples dataset~\cite{tnt} to evaluate the impact of voxel representation and SDF mapping. The results are presented in Tab. \ref{tab:app-ablation}. It can be observed that utilizing voxel representation significantly improves the PSNR for NVS tasks and reduces memory consumption dramatically compared to naive 3DGS setup. Although there is a slight decrease in the geometric quality of surface reconstruction, we consider this trade-off acceptable. 

\begin{table}[]
\centering
\small
\setlength{\tabcolsep}{3pt} 
\caption{Further ablation study on voxel Gaussian representation and SDF mapping. w/o voxel: We eliminate the using of MLPs and voxel grid, w/o sdf: we directly use linear assumption between opacity function and SDF function.}
\label{tab:app-ablation}
\begin{tabular}{ccccccc}
\toprule
\textbf{Ablation} & PSNR  & F1   & Mem (GB)       & Storage (MB) & Training Time & Meshing Time \\ \midrule
Ours              & 26.31         & 0.36         & $\sim$ 9 G          & 90 M          & $\sim$ 1.5 h       & $\sim$ 15 min     \\
w/o voxel         & 23.60 (-2.71) & 0.39 (+0.03) & $\sim$ 16 G ($\times$ 1.6) & 467 M ($\times$ 5.2)  & $\sim$ 1.4 h       & $\sim$ 15 min     \\
w/o sdf           & 26.31         & 0.30 (-0.06) & $\sim$ 9 G          & 90 M          & $\sim$ 1.5 h       & $\sim$15 min     \\ \bottomrule
\end{tabular}
\vspace{-15pt}
\end{table}

\subsection{Limitation}
\label{Limitation}
Implicit methods, such as those based on NeRF \cite{nerf,wang2023neus,li2023neuralangelo}, typically utilize a global fitting approach for SDF, which allows them to fully leverage the universal approximation capabilities of MLPs. This is advantageous even in areas with sparse viewpoints. However, our current method employs a local line-of-sight-based SDF fitting, a compromise made to adapt to the 3DGS rendering style. This means that regions not covered by the training viewpoints lack fitting capability, resulting in uneven surfaces.

In addition, While our method advances 3D surface reconstruction in open scenes, it faces challenges with dynamic objects and the decoupling of distant and near views, sometimes misrepresenting the sky as a surface enveloping the model. The lack of sufficient prior knowledge for optimizing complex scenes also poses limitations.

\section{Conclusion}
\label{Conclusion}
This paper introduces GVKF, combining Gaussian splatting's rapid rasterization with the efficiency of implicit expressions to enhance reconstruction quality and speed significantly. By employing a voxelized implicit representation of 3DGS, GVKF retains the expressive power of explicit Gaussian maps while managing them effectively. We have explored the relationship between Gaussian splatting's alpha blending and traditional volume rendering, developing a GS-based method to represent continuous scene opacity density fields through kernel regression, addressing 3DGS's limitations in continuous scene representation.

Experimental results demonstrate GVKF's effectiveness in open scenes, showing notable improvements in reconstruction accuracy, real-time rendering speeds, and reductions in storage and memory usage. These advancements support applications in fields like autonomous driving and virtual reality, pushing forward surface reconstruction technology.


\section*{Acknowledgements}
This research is supported by Guangzhou-HKUST(GZ) Joint Funding Program (Grant No.2023A03J0008), Education Bureau of Guangzhou Municipality, and Guangzhou Quwan Network Technology Co., Ltd.

\clearpage
\bibliographystyle{plain}
\bibliography{ref}

\begin{thebibliography}{10}

\bibitem{barron2021mipnerf}
Jonathan~T. Barron, Ben Mildenhall, Matthew Tancik, Peter Hedman, Ricardo Martin-Brualla, and Pratul~P. Srinivasan.
\newblock Mip-nerf: A multiscale representation for anti-aliasing neural radiance fields, 2021.

\bibitem{barron2023zipnerf}
Jonathan~T. Barron, Ben Mildenhall, Dor Verbin, Pratul~P. Srinivasan, and Peter Hedman.
\newblock Zip-nerf: Anti-aliased grid-based neural radiance fields.
\newblock {\em ICCV}, 2023.

\bibitem{chen2022tensorf}
Anpei Chen, Zexiang Xu, Andreas Geiger, Jingyi Yu, and Hao Su.
\newblock Tensorf: Tensorial radiance fields, 2022.

\bibitem{chen2024periodic}
Yurui Chen, Chun Gu, Junzhe Jiang, Xiatian Zhu, and Li~Zhang.
\newblock Periodic vibration gaussian: Dynamic urban scene reconstruction and real-time rendering, 2024.

\bibitem{chen2023neurbf}
Zhang Chen, Zhong Li, Liangchen Song, Lele Chen, Jingyi Yu, Junsong Yuan, and Yi~Xu.
\newblock Neurbf: A neural fields representation with adaptive radial basis functions, 2023.

\bibitem{ucnerf}
Kai Cheng, Xiaoxiao Long, Wei Yin, Jin Wang, Zhiqiang Wu, Yuexin Ma, Kaixuan Wang, Xiaozhi Chen, and Xuejin Chen.
\newblock Uc-nerf: Neural radiance field for under-calibrated multi-view cameras in autonomous driving, 2023.

\bibitem{darmon2022improving}
François Darmon, Bénédicte Bascle, Jean-Clément Devaux, Pascal Monasse, and Mathieu Aubry.
\newblock Improving neural implicit surfaces geometry with patch warping, 2022.

\bibitem{mt1}
Akio Doi and Akio Koide.
\newblock An efficient method of triangulating equi-valued surfaces by using tetrahedral cells.
\newblock {\em IEICE TRANSACTIONS on Information and Systems}, 74(1):214--224, 1991.

\bibitem{fu2022geoneus}
Qiancheng Fu, Qingshan Xu, Yew-Soon Ong, and Wenbing Tao.
\newblock Geo-neus: Geometry-consistent neural implicit surfaces learning for multi-view reconstruction, 2022.

\bibitem{intersec1}
Jian Gao, Chun Gu, Youtian Lin, Hao Zhu, Xun Cao, Li~Zhang, and Yao Yao.
\newblock Relightable 3d gaussian: Real-time point cloud relighting with brdf decomposition and ray tracing.
\newblock {\em arXiv preprint arXiv:2311.16043}, 2023.

\bibitem{guo2023streetsurf}
Jianfei Guo, Nianchen Deng, Xinyang Li, Yeqi Bai, Botian Shi, Chiyu Wang, Chenjing Ding, Dongliang Wang, and Yikang Li.
\newblock Streetsurf: Extending multi-view implicit surface reconstruction to street views.
\newblock {\em arXiv preprint arXiv:2306.04988}, 2023.

\bibitem{guédon2023sugar}
Antoine Guédon and Vincent Lepetit.
\newblock Sugar: Surface-aligned gaussian splatting for efficient 3d mesh reconstruction and high-quality mesh rendering, 2023.

\bibitem{deepblend}
Peter Hedman, Julien Philip, True Price, Jan-Michael Frahm, George Drettakis, and Gabriel Brostow.
\newblock Deep blending for free-viewpoint image-based rendering.
\newblock {\em ACM Trans. Graph.}, 37(6), dec 2018.

\bibitem{huang20242d}
Binbin Huang, Zehao Yu, Anpei Chen, Andreas Geiger, and Shenghua Gao.
\newblock 2d gaussian splatting for geometrically accurate radiance fields, 2024.

\bibitem{kerbl3Dgaussians}
Bernhard Kerbl, Georgios Kopanas, Thomas Leimkühler, and George Drettakis.
\newblock 3d gaussian splatting for real-time radiance field rendering, 2023.

\bibitem{intersec2}
Leonid Keselman and Martial Hebert.
\newblock Approximate differentiable rendering with algebraic surfaces.
\newblock In {\em European Conference on Computer Vision}, pages 596--614. Springer, 2022.

\bibitem{tnt}
Arno Knapitsch, Jaesik Park, Qian-Yi Zhou, and Vladlen Koltun.
\newblock Tanks and temples: benchmarking large-scale scene reconstruction.
\newblock {\em ACM Transactions on Graphics}, 36:1--13, 07 2017.

\bibitem{Kopanas_2022}
Georgios Kopanas, Thomas Leimkühler, Gilles Rainer, Clément Jambon, and George Drettakis.
\newblock Neural point catacaustics for novel-view synthesis of reflections.
\newblock {\em ACM Transactions on Graphics}, 41(6):1–15, November 2022.

\bibitem{kulhanek2023tetranerf}
Jonas Kulhanek and Torsten Sattler.
\newblock Tetra-nerf: Representing neural radiance fields using tetrahedra, 2023.

\bibitem{li2023neuralangelo}
Zhaoshuo Li, Thomas Müller, Alex Evans, Russell~H. Taylor, Mathias Unberath, Ming-Yu Liu, and Chen-Hsuan Lin.
\newblock Neuralangelo: High-fidelity neural surface reconstruction, 2023.

\bibitem{lin2024vastgaussian}
Jiaqi Lin, Zhihao Li, Xiao Tang, Jianzhuang Liu, Shiyong Liu, Jiayue Liu, Yangdi Lu, Xiaofei Wu, Songcen Xu, Youliang Yan, and Wenming Yang.
\newblock Vastgaussian: Vast 3d gaussians for large scene reconstruction, 2024.

\bibitem{liu2021neural}
Lingjie Liu, Jiatao Gu, Kyaw~Zaw Lin, Tat-Seng Chua, and Christian Theobalt.
\newblock Neural sparse voxel fields, 2021.

\bibitem{matchingcubes}
William~E. Lorensen and Harvey~E. Cline.
\newblock Marching cubes: A high resolution 3d surface construction algorithm.
\newblock In {\em Proceedings of the 14th Annual Conference on Computer Graphics and Interactive Techniques}, SIGGRAPH '87, page 163–169, New York, NY, USA, 1987. Association for Computing Machinery.

\bibitem{mcal}
William~E Lorensen and Harvey~E Cline.
\newblock Marching cubes: A high resolution 3d surface construction algorithm.
\newblock In {\em Seminal graphics: pioneering efforts that shaped the field}, pages 347--353. 1998.

\bibitem{lu2023scaffoldgs}
Tao Lu, Mulin Yu, Linning Xu, Yuanbo Xiangli, Limin Wang, Dahua Lin, and Bo~Dai.
\newblock Scaffold-gs: Structured 3d gaussians for view-adaptive rendering, 2023.

\bibitem{mi2023switchnerf}
Zhenxing Mi and Dan Xu.
\newblock Switch-nerf: Learning scene decomposition with mixture of experts for large-scale neural radiance fields.
\newblock In {\em International Conference on Learning Representations (ICLR)}, 2023.

\bibitem{nerf}
Ben Mildenhall, Pratul~P. Srinivasan, Matthew Tancik, Jonathan~T. Barron, Ravi Ramamoorthi, and Ren Ng.
\newblock Nerf: Representing scenes as neural radiance fields for view synthesis.
\newblock {\em Commun. ACM}, 65(1):99–106, dec 2021.

\bibitem{instantngp}
Thomas Müller, Alex Evans, Christoph Schied, and Alexander Keller.
\newblock Instant neural graphics primitives with a multiresolution hash encoding.
\newblock {\em ACM Transactions on Graphics}, page 1–15, Jul 2022.

\bibitem{oechsle2021unisurf}
Michael Oechsle, Songyou Peng, and Andreas Geiger.
\newblock Unisurf: Unifying neural implicit surfaces and radiance fields for multi-view reconstruction, 2021.

\bibitem{Surfels}
Hanspeter Pfister, Matthias Zwicker, Jeroen van Baar, and Markus Gross.
\newblock Surfels: surface elements as rendering primitives.
\newblock In {\em Proceedings of the 27th Annual Conference on Computer Graphics and Interactive Techniques}, SIGGRAPH '00, page 335–342, USA, 2000. ACM Press/Addison-Wesley Publishing Co.

\bibitem{merf}
Christian Reiser, Rick Szeliski, Dor Verbin, Pratul Srinivasan, Ben Mildenhall, Andreas Geiger, Jon Barron, and Peter Hedman.
\newblock Merf: Memory-efficient radiance fields for real-time view synthesis in unbounded scenes.
\newblock {\em ACM Trans. Graph.}, 42(4), jul 2023.

\bibitem{ren2024octreegs}
Kerui Ren, Lihan Jiang, Tao Lu, Mulin Yu, Linning Xu, Zhangkai Ni, and Bo~Dai.
\newblock Octree-gs: Towards consistent real-time rendering with lod-structured 3d gaussians, 2024.

\bibitem{mt2}
Tianchang Shen, Jun Gao, Kangxue Yin, Ming-Yu Liu, and Sanja Fidler.
\newblock Deep marching tetrahedra: a hybrid representation for high-resolution 3d shape synthesis.
\newblock {\em Advances in Neural Information Processing Systems}, 34:6087--6101, 2021.

\bibitem{sun2022direct}
Cheng Sun, Min Sun, and Hwann-Tzong Chen.
\newblock Direct voxel grid optimization: Super-fast convergence for radiance fields reconstruction, 2022.

\bibitem{Sun}
Pei Sun, Henrik Kretzschmar, Xerxes Dotiwalla, Aurelien Chouard, Vijaysai Patnaik, Paul Tsui, James Guo, Yin Zhou, Yuning Chai, Benjamin Caine, Vijay Vasudevan, Wei Han, Jiquan Ngiam, Hang Zhao, Aleksei Timofeev, Scott Ettinger, Maxim Krivokon, Amy Gao, Aditya Joshi, Yu~Zhang, Jonathon Shlens, Zhifeng Chen, and Dragomir Anguelov.
\newblock Scalability in perception for autonomous driving: Waymo open dataset.
\newblock In {\em 2020 IEEE/CVF Conference on Computer Vision and Pattern Recognition (CVPR)}, Jun 2020.

\bibitem{tancik2022blocknerf}
Matthew Tancik, Vincent Casser, Xinchen Yan, Sabeek Pradhan, Ben Mildenhall, Pratul Srinivasan, Jonathan~T. Barron, and Henrik Kretzschmar.
\newblock {Block-NeRF}: Scalable large scene neural view synthesis.
\newblock {\em arXiv}, 2022.

\bibitem{wang2023neus}
Peng Wang, Lingjie Liu, Yuan Liu, Christian Theobalt, Taku Komura, and Wenping Wang.
\newblock Neus: Learning neural implicit surfaces by volume rendering for multi-view reconstruction, 2023.

\bibitem{wang2023f2nerf}
Peng Wang, Yuan Liu, Zhaoxi Chen, Lingjie Liu, Ziwei Liu, Taku Komura, Christian Theobalt, and Wenping Wang.
\newblock F$^{2}$-nerf: Fast neural radiance field training with free camera trajectories, 2023.

\bibitem{wang2023neus2}
Yiming Wang, Qin Han, Marc Habermann, Kostas Daniilidis, Christian Theobalt, and Lingjie Liu.
\newblock Neus2: Fast learning of neural implicit surfaces for multi-view reconstruction, 2023.

\bibitem{snerf}
Ziyang Xie, Junge Zhang, Wenye Li, Feihu Zhang, and Li~Zhang.
\newblock S-nerf: Neural radiance fields for street views.
\newblock Mar 2023.

\bibitem{yan2023streetgaussians}
Yunzhi Yan, Haotong Lin, Chenxu Zhou, Weijie Wang, Haiyang Sun, Kun Zhan, Xianpeng Lang, Xiaowei Zhou, and Sida Peng.
\newblock Street gaussians for modeling dynamic urban scenes.
\newblock 2023.

\bibitem{yang2022neumesh}
Bangbang Yang, Chong Bao, Junyi Zeng, Hujun Bao, Yinda Zhang, Zhaopeng Cui, and Guofeng Zhang.
\newblock Neumesh: Learning disentangled neural mesh-based implicit field for geometry and texture editing, 2022.

\bibitem{NEURIPS2021_25e2a30f}
Lior Yariv, Jiatao Gu, Yoni Kasten, and Yaron Lipman.
\newblock Volume rendering of neural implicit surfaces.
\newblock In M.~Ranzato, A.~Beygelzimer, Y.~Dauphin, P.S. Liang, and J.~Wortman Vaughan, editors, {\em Advances in Neural Information Processing Systems}, volume~34, pages 4805--4815. Curran Associates, Inc., 2021.

\bibitem{yariv2023bakedsdf}
Lior Yariv, Peter Hedman, Christian Reiser, Dor Verbin, Pratul~P. Srinivasan, Richard Szeliski, Jonathan~T. Barron, and Ben Mildenhall.
\newblock Bakedsdf: Meshing neural sdfs for real-time view synthesis, 2023.

\bibitem{yu2021plenoctrees}
Alex Yu, Ruilong Li, Matthew Tancik, Hao Li, Ren Ng, and Angjoo Kanazawa.
\newblock Plenoctrees for real-time rendering of neural radiance fields, 2021.

\bibitem{yu2024gsdf}
Mulin Yu, Tao Lu, Linning Xu, Lihan Jiang, Yuanbo Xiangli, and Bo~Dai.
\newblock Gsdf: 3dgs meets sdf for improved rendering and reconstruction, 2024.

\bibitem{yu2024gaussian}
Zehao Yu, Torsten Sattler, and Andreas Geiger.
\newblock Gaussian opacity fields: Efficient and compact surface reconstruction in unbounded scenes, 2024.

\bibitem{zhang2021ners}
Jason~Y. Zhang, Gengshan Yang, Shubham Tulsiani, and Deva Ramanan.
\newblock Ners: Neural reflectance surfaces for sparse-view 3d reconstruction in the wild, 2021.

\bibitem{964490}
M.~Zwicker, H.~Pfister, J.~van Baar, and M.~Gross.
\newblock Ewa volume splatting.
\newblock In {\em Proceedings Visualization, 2001. VIS '01.}, pages 29--538, 2001.

\end{thebibliography}

\clearpage
\appendix
\section{Appendix / Supplemental Material}
\subsection{Ray-Gaussian Intersection}
\label{appendix:raygaussian}

\begin{wrapfigure}{r}{0.45\textwidth} 
    \centering
    \includegraphics[width=0.45\textwidth]{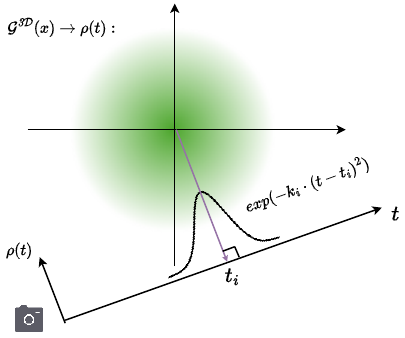} 
    \caption{Ray-Gaussian Intersection in local 3DGS coordinate.}
    \label{fig:app-rayga}
\end{wrapfigure}
Based on Eq. \ref{origin3dgs} in the main text, the influence of 3DGS in camera space on a one-dimensional ray can be expressed as follows:
\begin{equation}
    \rho(t) = \exp(-\frac{1}{2}(vt-p)\Sigma^{-1}(vt-p))
\end{equation}
Here, $v$ represents the unit vector of the ray direction. This formula converts the three-dimensional influence of 3DGS into a one-dimensional function along a specific camera ray, which is a one-dimensional Gaussian function. Fig. \ref{fig:app-rayga}  demonstrates the relationship of this transform. For ease of notation, we express it as:
\begin{equation}
    \rho(t) = \exp(-k_i \cdot (t-t_i)^2)
\end{equation}
where $t_i$ denotes the point along the ray where 3DGS has the maximum impact, also known as the "ray-Gaussian intersection," which can be analytically given by:
\begin{equation}
    t_i = \frac{p^T \Sigma^{-1} v}{v^T \Sigma^{-1} v}
\end{equation}
More proof details can be found in "Approximate Differentiable Rendering with Algebraic Surfaces." \cite{intersec2}

\subsection{More Implementation Details}
\label{appdix:moredetails}
As demonstrated in the ablation experiments, to balance quality and speed, we chose to downsample the initial Gaussian point cloud using a voxel size of 0.01. Within each voxel, the dimension of $\mathcal{F}$ is set to 32, and it stores the relative coordinates of 10 Gaussian points, indicating that the maximum number of Gaussians generated per voxel grid is 10. Gaussians with an opacity less than 0 will be hidden during each iteration. In each scene, all voxel grids share a total of four MLPs, which decode different Gaussian attributes from the corresponding voxels.

Regarding voxel registration, the gradient threshold is empirically set to \( 2 \times 10^{-4} \), meaning that voxel grids with an average gradient exceeding this value after each iteration will be subdivided using an octree method. The maximum recursion depth is set to 3 to control the number of Gaussians in the scene, ensuring it does not exceed a certain threshold. Voxel evaluation is performed every 500 iterations to determine which voxels should be subdivided or reclaimed. For other settings, we strive to remain consistent with the original 3DGS settings.

\begin{figure}

\centering
  \includegraphics[width=1\linewidth]{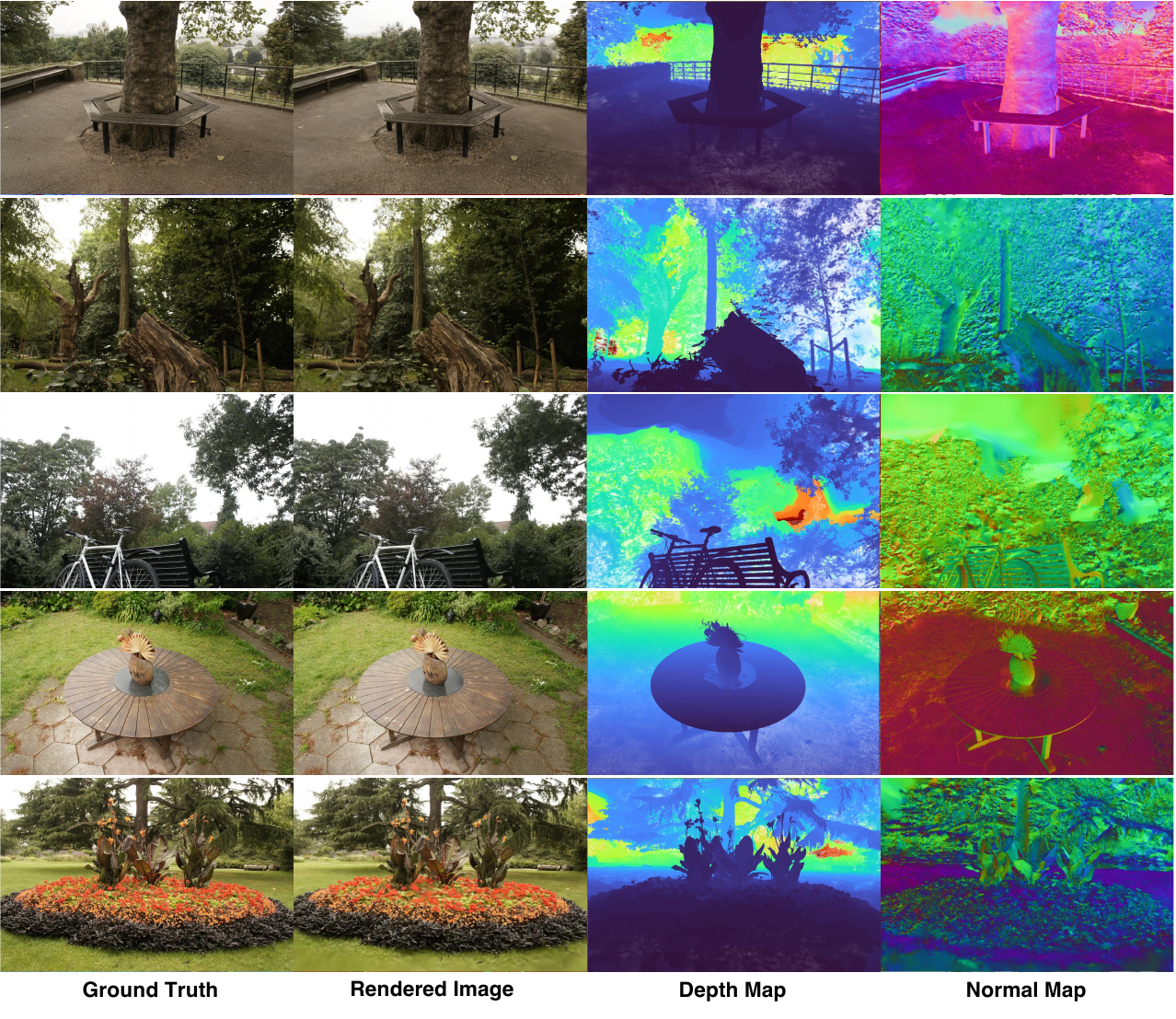}

  \caption{Additional experimental results on the Mip-NeRF360 dataset \cite{barron2021mipnerf}. From left to right: Ground Truth, Novel View Synthesis, Rendered Depth Map, and Normal Map.}
  \label{fig:mip_app}

\end{figure}

\begin{figure}

\centering
  \includegraphics[width=1\linewidth]{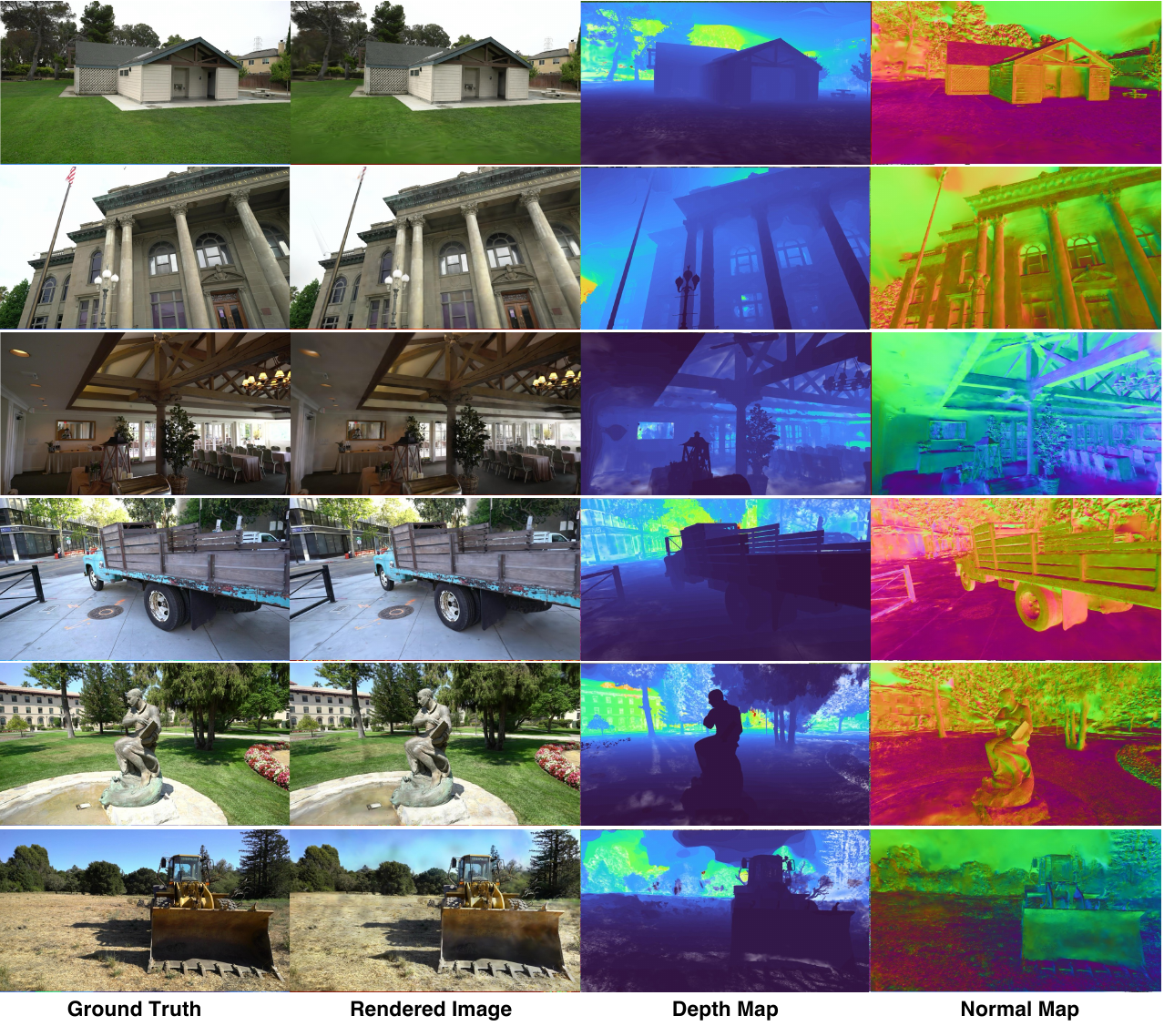}

  \caption{Additional experimental results on the Tanks and Temples dataset \cite{tnt}. From left to right: Ground Truth, Novel View Synthesis, Rendered Depth Map, and Normal Map.}
  \label{fig:tnt_app}

\end{figure}

\begin{figure} 

\centering
\includegraphics[width=1\textwidth]{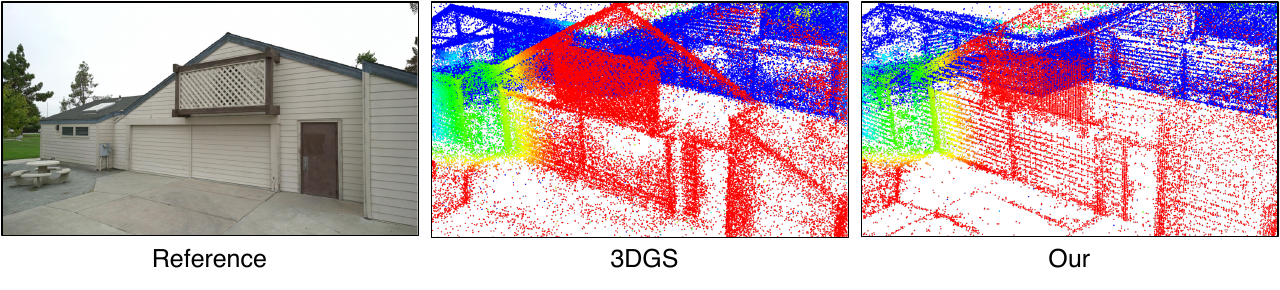} 

\caption{GVKF Gaussian point visualization compared to traditional Gaussian method.}
\label{fig:appdix-indoor}

\end{figure}

\subsection{More Results}
\label{appendix:moreresults}
Our method focuses on the challenging task of open scene reconstruction. Here, we provide a comprehensive quantitative comparison with other related methods on the Mip360 dataset, as shown in Table \ref{tab:mip_app}. Additionally, we have included more experimental results on the Mip360 and Tank and Temple datasets, as shown in Figures \ref{fig:mip_app} and \ref{fig:tnt_app}. For more qualitative results, please visit the project page.

\textbf{Discussion on indoor scene.} We observe that current methods based on 3DGS perform adequately for indoor scenes, where there is typically 360-degree viewpoint coverage. However, they underperform in outdoor scenes due to limited viewpoint coverage. Heuristic splitting and pruning strategies in original 3DGS tend to fit the training viewpoints rather than distributing evenly across the space. This leads to poorer novel view synthesis results in outdoor environments. As illustrated in Fig. \ref{fig:appdix-indoor}, without a voxel grid, heuristic Gaussian growth strategies result in uneven spatial distribution of GS, sometimes even creating holes. Conversely, using voxel grids to constrain Gaussians allows for efficient management of their spatial distribution, supporting better novel view synthesis.

\begin{figure} 
\centering
\includegraphics[width=1\textwidth]{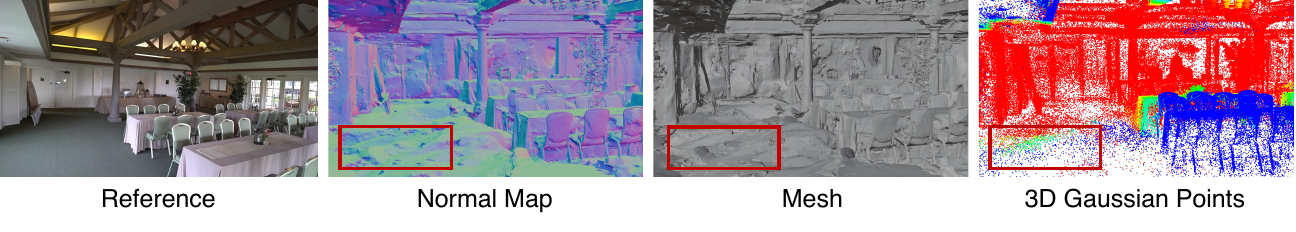} 
\vspace{-10pt}
\caption{Failure case. The sparse view area with less Gaussians tends to appear uneven surface.}
\label{fig:limit}
\end{figure}

\begin{table}
\centering
\small
\setlength{\tabcolsep}{3pt} 
\caption{Quantitative evaluation on the Mip-NeRF 360~\cite{barron2021mipnerf}. All scene dataset is presented.}
\label{tab:mip_app}
\begin{tabular}{@{}cccccccccccc@{}}
\toprule
\multirow{2}{*}{\textbf{Method}} & \multicolumn{3}{c}{\textbf{Outdoor Scene}} & \multicolumn{3}{c}{\textbf{Indoor Scene}} & \multicolumn{3}{c}{\textbf{Average}} \\ \cmidrule(lr){2-4} \cmidrule(lr){5-7} \cmidrule(lr){8-10}
 & \textbf{PSNR} $\uparrow$ & \textbf{SSIM} $\uparrow$ & \textbf{LPIPS} $\downarrow$& \textbf{PSNR} $\uparrow$& \textbf{SSIM} $\uparrow$& \textbf{LPIPS} $\downarrow$& \textbf{PSNR} $\uparrow$& \textbf{SSIM} $\uparrow$& \textbf{LPIPS} $\downarrow$\\ \midrule
NeRF & 21.46 & 0.458 & 0.515 & 26.84 & 0.79 & 0.37 & 23.85 & 0.61 & 0.45 \\
Deep Blending & 21.54 & 0.524 & 0.364 & 26.4 & 0.844 & 0.261 & 23.70 & 0.67 & 0.32 \\
Instant NGP & 22.9 & 0.566 & 0.371 & 29.15 & 0.88 & 0.216 & 25.68 & 0.72 & 0.30 \\
MERF & 23.19 & 0.616 & 0.343 & 27.8 & 0.855 & 0.271 & 25.24 & 0.72 & 0.31 \\
MipNeRF360 & 24.47 & 0.691 & 0.283 & \textbf{31.72} & 0.917 & \textbf{0.18} & 27.69 & 0.79 & 0.24 \\
BakedSDF & 22.47 & 0.585 & 0.349 & 27.06 & 0.836 & 0.258 & 24.51 & 0.70 & 0.30 \\
Mobile-NeRF & 21.95 & 0.470 & 0.470 & 12.19 & 0.26 & 0.26 & 17.07 & 0.36 & 0.37 \\
3DGS & 24.24 & 0.705 & 0.283 & 30.99 & \textbf{0.926} & 0.199 & 27.24 & 0.80 & 0.25 \\
SuGaR & 22.76 & 0.631 & 0.349 & 29.44 & 0.911 & 0.216 & 25.73 & 0.76 & 0.29 \\
2DGS & 24.33 & 0.709 & 0.284 & 30.39 & 0.924 & 0.182 & 27.02 & 0.80 & 0.24 \\
GVKF (Ours) & \textbf{25.47} & \textbf{0.757} & \textbf{0.240} & 30 & 0.915 & 0.2 & \textbf{27.48} & \textbf{0.83} & \textbf{0.22} \\ \bottomrule
\end{tabular}
\end{table}

\textbf{Failure Case.} As shown in Fig. \ref{fig:limit}, in areas with sparse viewpoint coverage, the distribution of 3DGS is sparse and irregular, which hinders the fitting of smooth planes. This sparsity compromises the integrity of the surface reconstruction, resulting in models that are geometrically inaccurate. Our method still struggles to effectively address these issues.

\begin{table}
\centering
\small
\setlength{\tabcolsep}{3pt} 
\caption{NVS and storage comparation to GOF on Mip-NeRF 360 Dataset \cite{barron2021mipnerf}.}
\label{tab:app-comptogof}
\begin{tabular}{ccccc}
\toprule
Mip-NeRF 360     & PSNR $\uparrow$  & SSIM $\uparrow$  & LPIPS $\downarrow$ & Storage $\downarrow$ \\ \midrule
GOF         & 24.53 & 0.733 & 0.245 & 649 M   \\
GVKF (ours) & \textbf{25.47} & \textbf{0.757 }& \textbf{0.240} & \textbf{68 M}    \\ \midrule
\end{tabular}
\end{table}

\begin{table}
\centering
\small
\setlength{\tabcolsep}{3pt} 
\caption{Mesh quality comparation (F1-score) to GOF on Tanks and Temples Dataset. \cite{tnt}}
\label{tab:app-comptogof2}
\begin{tabular}{cccccccc}
\toprule
TanT & Barn & Caterpillar & Courthouse & Ignatius & Meetingroom & Truck & Mean \\ \midrule
GOF  & \textbf{0.51} & \textbf{0.41   }     & \textbf{0.28 }      & \textbf{0.68}     & \textbf{0.28 }       &\textbf{ 0.59 } &\textbf{ 0.46} \\
GVKF (ours) & 0.40 & 0.34        & 0.25       & 0.51     & 0.23        & 0.40  & 0.36 \\ \bottomrule
\end{tabular}
\end{table}

\subsection{Comparation to Gaussian Opacity Field}
\label{appendix:comptogof}
The similar rendering equation is firstly proposed by GOF \cite{yu2024gaussian}, while this work provides in-depth analysis of the relationship among this rendering strategy, volume rendering and Gaussian alpha blending. Different from GOF, our scene representation is implicit, addressing the common issue of high memory consumption faced by 3D Gaussian splatting. Additionally, we developed a mapping function from opacity to SDF to alleviate the influence of directly linear transform between these fields.

As shown in Tab. \ref{tab:app-comptogof}, GOF uses explicit Gaussian management, still faces high storage consumption issues, making training large scenes on a single card challenging. Our method achieves better novel view synthesis results with less storage usage. However, as shown in Tab. \ref{tab:app-comptogof2}, our current implementation has some geometric precision gaps compared to GOF, the potential reasons may include:
\begin{itemize}
    \item GOF's iterative optimization extraction method achieves more precise isosurfaces than ours. (May require long meshing time $\sim 2$ h)
    \item Further adaptation of regularization term to voxel grids might be needed in our method to improve mesh quality.
\end{itemize}

\clearpage
\newpage
\section*{NeurIPS Paper Checklist}

The checklist is designed to encourage best practices for responsible machine learning research, addressing issues of reproducibility, transparency, research ethics, and societal impact. Do not remove the checklist: {\bf The papers not including the checklist will be desk rejected.} The checklist should follow the references and follow the (optional) supplemental material.  The checklist does NOT count towards the page
limit. 

Please read the checklist guidelines carefully for information on how to answer these questions. For each question in the checklist:
\begin{itemize}
    \item You should answer \answerYes{}, \answerNo{}, or \answerNA{}.
    \item \answerNA{} means either that the question is Not Applicable for that particular paper or the relevant information is Not Available.
    \item Please provide a short (1–2 sentence) justification right after your answer (even for NA). 
\end{itemize}

{\bf The checklist answers are an integral part of your paper submission.} They are visible to the reviewers, area chairs, senior area chairs, and ethics reviewers. You will be asked to also include it (after eventual revisions) with the final version of your paper, and its final version will be published with the paper.

The reviewers of your paper will be asked to use the checklist as one of the factors in their evaluation. While "\answerYes{}" is generally preferable to "\answerNo{}", it is perfectly acceptable to answer "\answerNo{}" provided a proper justification is given (e.g., "error bars are not reported because it would be too computationally expensive" or "we were unable to find the license for the dataset we used"). In general, answering "\answerNo{}" or "\answerNA{}" is not grounds for rejection. While the questions are phrased in a binary way, we acknowledge that the true answer is often more nuanced, so please just use your best judgment and write a justification to elaborate. All supporting evidence can appear either in the main paper or the supplemental material, provided in appendix. If you answer \answerYes{} to a question, in the justification please point to the section(s) where related material for the question can be found.

IMPORTANT, please:
\begin{itemize}
    \item {\bf Delete this instruction block, but keep the section heading ``NeurIPS paper checklist"},
    \item  {\bf Keep the checklist subsection headings, questions/answers and guidelines below.}
    \item {\bf Do not modify the questions and only use the provided macros for your answers}.
\end{itemize}


\begin{enumerate}

\item {\bf Claims}
    \item[] Question: Do the main claims made in the abstract and introduction accurately reflect the paper's contributions and scope?
    \item[] Answer: \answerYes{} 
    \item[] Justification: The abstract and introduction clearly outline the contributions of the paper, including the development of a novel method for open scene surface reconstruction that combines the strengths of explicit and implicit approaches, and its application to large-scale scene reconstruction tasks (Sec. \ref{Introduction}).

\item {\bf Limitations}
    \item[] Question: Does the paper discuss the limitations of the work performed by the authors?
    \item[] Answer: \answerYes{} 
    \item[] Justification:  "Limitations" section is included in the paper, discussing the constraints of the proposed method (Sec. \ref{Limitation})

\item {\bf Theory Assumptions and Proofs}
    \item[] Question: For each theoretical result, does the paper provide the full set of assumptions and a complete (and correct) proof?
    \item[] Answer: \answerYes{} 
    \item[] Justification: Please see Sec. \ref{Methods} for relevant proofs and assumptions.

    \item {\bf Experimental Result Reproducibility}
    \item[] Question: Does the paper fully disclose all the information needed to reproduce the main experimental results of the paper to the extent that it affects the main claims and/or conclusions of the paper (regardless of whether the code and data are provided or not)?
    \item[] Answer: \answerYes{} 
    \item[] Justification: The paper provides detailed information on the datasets used, the experimental setup and other relevant details necessary to reproduce the results. (Sections \ref{Experiments})

\item {\bf Open access to data and code}
    \item[] Question: Does the paper provide open access to the data and code, with sufficient instructions to faithfully reproduce the main experimental results, as described in supplemental material?
    \item[] Answer: \answerYes{} 
    \item[] Justification: The code and data are available in a public repository.

\item {\bf Experimental Setting/Details}
    \item[] Question: Does the paper specify all the training and test details (e.g., data splits, hyperparameters, how they were chosen, type of optimizer, etc.) necessary to understand the results?
    \item[] Answer: \answerYes{} 
    \item[] Justification: All relevant training and testing details, including data splits, hyperparameters, optimizer types, and selection criteria, are clearly specified in the paper. (Sec. \ref{Experiments})

\item {\bf Experiment Statistical Significance}
    \item[] Question: Does the paper report error bars suitably and correctly defined or other appropriate information about the statistical significance of the experiments?
    \item[] Answer: \answerNo{} 
    \item[] Justification: The paper reports PSNR, SSIM, F1 scores, LPIPS,and C-D values, which are commonly used as a measure of performance in image processing experiments. This approach is standard in the field and is sufficient to convey the performance of the methods under investigation. 

\item {\bf Experiments Compute Resources}
    \item[] Question: For each experiment, does the paper provide sufficient information on the computer resources (type of compute workers, memory, time of execution) needed to reproduce the experiments?
    \item[] Answer: \answerYes{} 
    \item[] Justification: The type of compute resources used, including GPU specifications, memory, and execution time for each experiment, is detailed in the paper. (Sec. \ref{Experiments})
    
\item {\bf Code Of Ethics}
    \item[] Question: Does the research conducted in the paper conform, in every respect, with the NeurIPS Code of Ethics \url{https://neurips.cc/public/EthicsGuidelines}?
    \item[] Answer: \answerYes{} 
    \item[] Justification: The research adheres to the NeurIPS Code of Ethics, with considerations for reproducibility, transparency, and societal impact addressed throughout the paper.

\item {\bf Broader Impacts}
    \item[] Question: Does the paper discuss both potential positive societal impacts and negative societal impacts of the work performed?
    \item[] Answer: \answerYes{} 
    \item[] Justification: The paper includes a section discussing the broader impacts of the proposed method, highlighting both potential positive applications and possible negative consequences. (Sec. \ref{Conclusion})
    
\item {\bf Safeguards}
    \item[] Question: Does the paper describe safeguards that have been put in place for responsible release of data or models that have a high risk for misuse (e.g., pretrained language models, image generators, or scraped datasets)?
    \item[] Answer: \answerNA{} 
    \item[] Justification: The paper does not release any data or models that are considered to have a high risk for misuse.

\item {\bf Licenses for existing assets}
    \item[] Question: Are the creators or original owners of assets (e.g., code, data, models), used in the paper, properly credited and are the license and terms of use explicitly mentioned and properly respected?
    \item[] Answer: \answerYes{} 
    \item[] Justification: All existing assets used in the paper are properly credited, and their licenses and terms of use are explicitly respected. (Sec. \ref{Experiments})

\item {\bf New Assets}
    \item[] Question: Are new assets introduced in the paper well documented and is the documentation provided alongside the assets?
    \item[] Answer: \answerNA{} 
    \item[] Justification: The paper does not introduce any new assets.

\item {\bf Crowdsourcing and Research with Human Subjects}
    \item[] Question: For crowdsourcing experiments and research with human subjects, does the paper include the full text of instructions given to participants and screenshots, if applicable, as well as details about compensation (if any)? 
    \item[] Answer: \answerNA{} 
    \item[] Justification: The paper does not involve crowdsourcing experiments or research with human subjects.

\item {\bf Institutional Review Board (IRB) Approvals or Equivalent for Research with Human Subjects}
    \item[] Question: Does the paper describe potential risks incurred by study participants, whether such risks were disclosed to the subjects, and whether Institutional Review Board (IRB) approvals (or an equivalent approval/review based on the requirements of your country or institution) were obtained?
    \item[] Answer: \answerNA{} 
    \item[] Justification: The paper does not involve research with human subjects.

\end{enumerate}

\clearpage




\end{document}